\documentclass[english]{article}
\usepackage[latin9]{inputenc}
\usepackage{mathrsfs}
\usepackage{amssymb}
\usepackage{graphicx}

\makeatletter
\@ifundefined{showcaptionsetup}{}{%
 \PassOptionsToPackage{caption=false}{subfig}}
\usepackage{subfig}
\makeatother

\usepackage{babel}
\usepackage{listings}

\begin{document}

\title{The Computational Theory of Intelligence: Information Entropy}

\author{Daniel Kovach}
\maketitle
\begin{abstract}
This paper presents an information theoretic approach to the concept
of intelligence in the computational sense. We introduce a probabilistic
framework from which computational intelligence is shown to be an
entropy minimizing process at the local level. Using this new scheme,
we develop a simple data driven clustering example and discuss its
applications.
\end{abstract}

\section{Introduction}

This paper attempts to introduce a computational approach to the study
of intelligence that the researcher has accumulated over years of
study. This approach takes into account data from psychology, neurology,
artificial intelligence, machine learning, and mathematics. 

Central to this framework is the fact that the goal of any intelligent
agent is to reduce the randomness in its environment in some meaningful
way. Of course, formal definitions in the context of this paper for
terms like \char`\"{}intelligence\char`\"{}, \char`\"{}environment\char`\"{},
and \char`\"{}agent\char`\"{} will follow.

The approach draws from multidisciplinary research and has many applications.
We will utilize the construct in discussions at the end of the paper.
Other applications will follow in future works. Implementations of
this framework can apply to many fields of study including general
artificial intelligence (GAI), machine learning, optimization, information
gathering, clustering, and big data, and extend outside of the applied
mathematics and computer science realm to even more areas including
sociology, psychology, and neurology, and even philosophy.

\subsection{Definitions}

One cannot begin a discussion about the philosophy of artificial intelligence
without a definition of the word ``intelligence'' in the first place.
With the panopoly of definitions available, it is understandable that
there may be some disagreement, but typically each school of thought
generally shares a common thread. The following are three different
definitions of \emph{intelligence} from respectable sources:
\begin{enumerate}
\item The aggregate or global capacity of the individual to act purposefully,
to think rationally, and to deal effectively with his environment\cite{we}.
\item A process that entails a set of skills of problem solving \textemdash{}
enabling the individual to resolve genuine problems or difficulties
that he or she encounters and, when appropriate, to create an effective
product \textemdash{} and must also entail the potential for finding
or creating problems \textemdash{} and thereby providing the foundation
for the acquisition of new knowledge\cite{ga}.
\item Goal-directed adaptive behavior\cite{ste}.
\end{enumerate}
Vernon's hierarchical model of intelligence from the 1950's \cite{aik},
and Hawkins' \emph{On Intelligence} in 2004 \cite{haw} are some other
great resources on this topic. Consider the following working definition
of this paper, with regard to information theory and computation:
\emph{computational intelligence }(CI) is an information processing
algorithm that 
\begin{enumerate}
\item Records data or \emph{events} into some type of store, or \emph{memory}.
\label{enu:records-events-into}
\item Draws from the events recorded in memory, to make assertions, or \emph{predictions}
about future events. \label{enu:It-then-draws}
\item Using the disparity between the predicted and events and the new incoming
events, the memory structure in step \ref{enu:records-events-into}
can be updated such that the predictions of step \ref{enu:It-then-draws}
are optimized. \label{enu:Using-the-disparity}
\end{enumerate}
The mapping in \ref{enu:Using-the-disparity} is called \emph{learning},
and is endemic to CI. Any entity that is facilitating the CI process
we will refer to as an \emph{agent}, in particular when the connotation
is that the entity is autonomous. The surrounding infrastructure that
encapsulates the agent together with the ensuing events is called
the \emph{environment}.

\subsection{Structure}

The paper is organized as follows. In section \ref{sec:Entropy} we
provide a brief summary of the concept of information entropy as it
is used for our purposes. In section \ref{sec:Intelligence}, we provide
a mathematical framework for intelligence and show discuss its relation
to entropy. Section \ref{sec:Global-Effects} discusses the global
ramifications of local entropy minimization. In section \ref{sec:Application}
we present a simple application of the framework to data analytics,
which is available for free download. Sections \ref{sec:Related-Work}
and \ref{sec:Conclusion} discuss relavent related research, and future
work.

\section{Entropy\label{sec:Entropy}}

A key concept of information theory is that of entropy, which amounts
to the uncertainty in a given random variable, \cite{ih}. It is essentially,
a measure of unpredictability (among other interpretations). The concept
of entropy is a much deeper principal of nature that penetrates to
the deepest core of physical reality and is central to physics and
cosmological models, \cite{sc,pen,hawk}.

\subsection{Mathematical Representation}

Although terms like \emph{Shannon entropy} are pervasive in the field
of information theory, it will be insightful to review the formulation
in our context. To arrive at the definition of entropy, we must first
recall what is meant by \emph{information content}. The information
content of a random variable, $X$ denoted $I\left[X\right]$, is
given by

\begin{equation}
I\left[X\right]=\log\left[\frac{1}{\mathbb{P}\left[X\right]}\right]=-\log\left[\mathbb{P}\left[X\right]\right]\label{eq:entropy}
\end{equation}

where $\mathbb{P}\left[X\right]$is the probability of $X$. The entropy
of $X$ , denoted $\mathbb{E}\left[X\right]$, is then defined as
the expecation value of the information content. 

\begin{equation}
\mathbb{E}\left[X\right]=\mathbb{E}\left[I\left[X\right]\right]=-\mathbb{E}\left[\log\left[\mathbb{P}\left[X\right]\right]\right]
\end{equation}

Expanding by the definition of the expectation value, we have

\begin{equation}
\mathbb{E}\left[X\right]=-\sum_{i=1}^{N}\mathbb{P}\left[x_{i}\right]\log\left[\mathbb{P}\left[x_{i}\right]\right]
\end{equation}

where $\left\{ x_{i}\right\} $is the set of possible values $X$
can take.

\subsubsection{Relationship of Shannon to Thermodynamic Entropy}

The concept of entropy is deeply rooted at the heart of physical reality.
It is a central concept in thermodynamics, governing everything from
chemical reactions to engines and refrigerators. The relationship
of entropy as it is known in information theory, however, is not mapped
so straightforwardly to its use in thermodynamics.

In statistical thermodynamics, the entropy $S$, of a system is given
by

\begin{equation}
S=-k_{b}\sum p_{i}\ln\left[p_{i}\right]
\end{equation}

where $p_{i}$ denote the probability of each \emph{microstate}, or
configuration of the system, and $k_{b}$ is the Boltzmann constant
which serves to map the value of the summation to physical reality
in terms of quantity and units.

The connection between the thermodynamic and information theoretic
versions of entropy relate to the information needed to detail the
exact state of the system, specifically, the amount of further Shannon
information needed to define the microscopic state of the system that
remains ambiguous when given its macroscopic definition in terms of
the variables of classical thermodynamics. The Boltzmann constant
serves as a constant of proportionality.

\subsubsection{Renyi Entropy}

We can extend the logic of the beginning of this section to a more
general formulation called the Renyi entropy of order $\alpha$, where
$\alpha\geq0$ and $\alpha\neq1$ defined as

\begin{equation}
\mathbb{H_{\alpha}}(X)=\frac{1}{1-\alpha}\log\left[\sum_{i=1}^{N}\mathbb{P}\left[x_{i}\right]^{\alpha}\right].\label{eq:renyi}
\end{equation}

Under this convention we can apply the concept of entropy more generally
to extend the utility of the concept to a variety of applications.
It is important to note that this formulation approches \ref{eq:entropy}
in the limit as $\alpha\rightarrow1$. Although the discussions of
this paper were inspired by Shannon entropy, we wish to present a
much more general definition and a bolder proposition.

\section{Intelligence: Definitions and Assumptions\label{sec:Intelligence}}

Consider the sets $\mathbb{S}$ and $\mathbb{O}$ and let $\mathbb{I}\in\mathscr{I}$
be the following mapping $\mathbb{I}:\mathbb{S}\rightarrow\mathbb{O}$.
The function $\mathbb{I}$ represents the intelligence process, a
member of $\mathscr{I}$, the set of all such functions. It maps input
from set $\mathbb{S}$ to $\mathbb{O}$. First, let 

\begin{equation}
\mathbb{I}_{t}\left[s^{i}\right]=o_{t}^{i}
\end{equation}

reflect the fact that $\mathbb{I}$ is mapping one element from $\mathbb{S}$
to one element in $\mathbb{O}$, each tagged by the identifier $i\in\mathbb{N}$,
which is bounded by the cardinality of the input set. The cardinality
of these two sets need not match, nor does the mapping between $\mathbb{I}$
need to be bijective, or even surjective. This is an iterative process,
as denoted by the index, $t$. Finally, let $\mathbb{\mathbb{O}}_{t}$
represent the collection of $o_{t}^{i}$.

Over time, the mapping should converge to the intended element, $o^{i}\in\mathbb{O}$,
as is reflected by

\begin{equation}
\mathbb{I}\left[s^{i}\right]\rightarrow o^{i},o^{i}\in\mathbb{O}.
\end{equation}

Introduce the function:

\begin{equation}
\mathbb{L}_{t}=f\left(\mathbb{O},\mathbb{\mathbb{O}}_{t}\right),
\end{equation}

which in practice is usually some type of error or fitness measuring
function. Assuming that $\mathbb{L}$ is continuous and differentiable,
let the updating mechanism at some particular $t$ for $\mathbb{I}$
be

\begin{equation}
\mathbb{I}_{t}=\mathbb{I}_{t-1}+\nabla\mathbb{L}_{t-1}.
\end{equation}

In other words, $\mathbb{I}$ iteratively updates itself with respect
to the gradient of some function, $\mathbb{L}$. Additionally, $\mathbb{L}$
must satisfy the following partial differential equation

\begin{equation}
\frac{\partial}{\partial t}\mathbb{L=}\rho(t)d\left(\mathbb{O}-\mathbb{\mathbb{O}}_{t}\right),\rho\mapsto\mathbb{R}
\end{equation}

where the function $d$ is some measure of the distance between $\mathbb{O}$
and $\mathbb{O}_{t}$, assuming such a function exists, and $\rho$
is called the \emph{learning rate}. Applications of this process to
abstract topological spaces where such a distance function is a commodity
is an open question. For this to qualify as an intelligence process,
we must have

\begin{equation}
\lim_{t\rightarrow\infty}d\left(\mathbb{O}-\mathbb{\mathbb{O}}_{t}\right)\rightarrow0.
\end{equation}

\subsection{Unsupervised and Supervised Learning\label{sub:Remarks}}

Some consideration should be given to the sets $\mathbb{S}$ and $\mathbb{O}$.
If $\mathbb{O}=P(\mathbb{S})$ where $P(\mathbb{S})$ is the power
set of $\mathbb{S}$, then we will say that the mapping $\mathbb{I}$
is an \emph{unsupervised} mapping. Otherwise, the mapping is \emph{supervised}.
The ramifications of this distinction is as follows. In supervised
learning, the agent is given two distinct sets and trained to form
a mapping between them \emph{explicitly}. With \emph{unsupervised
learning}, the agent is tasked with learning subtle relationships
in a single data set or, put more succinctly, to develop the mapping
between $\mathbb{S}$ and its power set discussed above \cite{jon,rus}.

\subsection{Overtraining}

Further, we should note that just because we have some function $\mathbb{I}:\mathbb{S}\rightarrow\mathbb{O}$
satisfying the definitions and assumptions of this section does not
mean that this mapping be necessarily \emph{meaningful}. After all,
we could make a completely arbitrary but consistent mapping via the
prescription above, and although this would satisfy all the definitions
and assumptions, it would be complete memorization on the part of
the agent. But this, in fact is exactly the definition of \emph{overtraining}
a common pitfall in the training stage of machine learning and about
which one must be very diligent to avoid.

\subsection{Entropy Minimization}

One final part of the framework remains, and that is to show that
entropy is minimized, as was stated at the beginning of this section.
To show that, we consider $\mathbb{I}$ as a \emph{probabilistic }mapping,
with

\begin{equation}
\mathbb{P}_{t}\left[s_{j}^{i}\right]=\mathbb{P}\left[\mathbb{I}_{t}\left[s_{j}^{i}\right]=o^{j}\right]
\end{equation}

indicating the probability that $\mathbb{I}$ maps $s_{j}^{i}\in\mathbb{S}$
to some $o^{j}\in\mathbb{O}$. From this, we can calculate the entropy
in the mapping from $\mathbb{S}$ to$\mathbb{O}$, at each iteration
$t$. If the projection $\mathbb{I}\left[s^{i}\right]$ has $N$ possible
outcomes, then the Shannon entropy of each$s^{i}\in\mathbb{S}$ is
given by

\begin{equation}
\mathbb{E}_{t}^{i}\left[s^{i}\right]=-\sum_{j=1}^{N}\mathbb{P}_{t}\left[s_{j}^{i}\right]\log\left[\mathbb{P}_{t}\left[s_{j}^{i}\right]\right].
\end{equation}

If $\left|\mathbb{S}\right|=M$, then the total entropy is simply
the sum of $\mathbb{E}_{t}^{i}\left[\mathbb{S}\right],i\in{1,2,...,N}$.
But for the purposes of standardization across varying cardinalities,
it is useful to speak of the \emph{normalized entropy}, 

\begin{equation}
\mathbb{E}_{t}\left[\mathbb{S}\right]=\frac{1}{M}\sum_{i=1}^{M}\mathbb{E}_{t}^{i}\left[s^{i}\right]\label{eq:weighted sum}
\end{equation}

As $t\rightarrow\infty$, the mapping from each $s^{i}$ to its corresponding
$o^{i}$ converges, and we have

\begin{equation}
\lim_{t\rightarrow\infty}\mathbb{P}_{t}\left[s^{i}\right]\rightarrow1,i\in{1,2,...,N}\label{eq:p=00003D1}
\end{equation}

and therefore

\begin{equation}
\lim_{t\rightarrow\infty}\mathbb{E}_{t}\rightarrow0
\end{equation}

Further, using the definition for Renyi entropy in \ref{eq:renyi}
for each $t$ and $i$

\begin{equation}
\mathbb{H}_{t,\alpha}^{i}\left[\mathbb{S}\right]=\frac{1}{1-\alpha}\log\left[\sum_{i=1}^{N}\mathbb{P}_{t}\left[s^{i}\right]^{\alpha}\right],i\in{1,2,...,N}
\end{equation}

To show that the Renyi entropy is also minimized, we can use an identity
involving the $p$-norm

\begin{equation}
\mathbb{H}_{t,\alpha}^{i}\left[\mathbb{S}\right]=\frac{\alpha}{1-\alpha}\log\left[\left\Vert \mathbb{P}_{t}\left[s^{i}\right]\right\Vert _{\alpha}\right],i\in{1,2,...,N}
\end{equation}

and show that the log function is maximized $t\rightarrow\infty$
for $\alpha>1$, and minimized for $\alpha<1$. The case where $\alpha\rightarrow1$
was shown above when we used the definition of Shannon entropy. To
continue, note that 

\begin{equation}
\sum\mathbb{P}_{t}\left[s^{i}\right]=1
\end{equation}

where the sum is taken over all possible states $o^{i}\in\mathbb{O}$
in the range of$\mathbb{I}_{t}\left[s^{i}\right]$. But from \ref{eq:p=00003D1},
we have

s
\begin{equation}
\left\Vert \mathbb{P}_{t}\left[s^{i}\right]\right\Vert _{\alpha}<1,\alpha>1
\end{equation}

for finite $t$ and thus the log function is minimized only as $t\rightarrow\infty$.
To show that the Renyi entropy is also minimized for $\alpha\in(0,1)$,
we repeat the above logic but note that the with the sign reversal
of $\frac{\alpha}{1-\alpha}$, we need to show that $\left\Vert \mathbb{P}_{t}\left[s^{i}\right]\right\Vert _{\alpha}$is
\emph{minimized} as $t\rightarrow\infty$.

\subsection{Entropic Self Organization\label{sub:Entropic-Self-Organization}}

In section \ref{sec:Intelligence} we talked about the definitions
of intelligence via the mapping $\mathbb{I}:\mathbb{S}\rightarrow\mathbb{O}$.
Here, we seek to apply the entropy minimization concept to $P(\mathbb{S})$
itself, rather than a mapping. Explicitly, let $\sigma\subset P(\mathbb{S})$ 

\begin{equation}
\sigma=\{\mathbf{s}\in P(\mathbb{S})\},\label{eq:sigma}
\end{equation}

where for every $s\in\mathbb{S}$, there is a unique $\mathbf{s}\in\sigma$
such that $s\in\mathbf{s}$. That is, every element of $\mathbb{S}$
has one and only one element of $\sigma$ containing it. The term
\emph{entropic self organization} refers to finding the $\Sigma\subset P(\mathbb{S})$
such that $\mathbb{H}_{\alpha}[\sigma]$ is minimized over all $\sigma$
satisfying \ref{eq:sigma}:

\begin{equation}
\mathbf{\Sigma}=\min\mathbb{H}_{\alpha}[\sigma].
\end{equation}

\section{Global Effects\label{sec:Global-Effects}}

In nature, whenever a system is taken from a state of higher entropy
to a state of lower entropy, there is always some amount of energy
involved in this transition, and an increase in the entropy of the
rest of the environment greater than or equal to that of the entropy
loss\cite{sc}. In other words, consider a system $S$ composed of
two subsystems, $s_{1}$ and $s_{2}$. Then 

\begin{equation}
S=s_{1}+s_{2}
\end{equation}

Now, consider that system in equilibrium at times $t=1$, and $t=2$,
denoted $S^{1}$ and $S^{2}$. Due to the second law of thermodynamics.

\begin{equation}
S^{2}\geq S^{1}
\end{equation}

and

\begin{equation}
s_{1}^{2}+s_{2}^{2}\geq s_{1}^{1}+s_{2}^{1}
\end{equation}

Now, suppose one of the subsystems, say, $s_{1}$decreases in entropy
by some amount, $\Delta s$ during the transition by time $t=2$,
i.e. $s_{1}^{2}=s_{1}^{1}-\Delta s$ Then what can be said of $s_{2}^{1}$,
the entropy of the rest of the system that

\begin{equation}
s_{2}^{2}\geq s_{2}^{1}+\Delta s
\end{equation}

So the entropy of the rest of the system has to increase by an amount
greater than or equal to the loss of entropy in $s_{1}$. This will
require some amount of energy, $\Delta E$.

While the second law of thermodynamics has been verified time and
again in virtually all areas of physics, few have extended it as a
more general principal in the context of information theory. In fact,
we will conclude this paper with a postulate about intelligence:
\begin{itemize}
\item \emph{Computational intelligence is a process that locally minimizes
and globally maximizes Renyi entropy.}
\end{itemize}
It should be stressed that although the above is necessary of intelligence,
it is not sufficient in the justification of an algorithm or process
as being intelligent.

\section{Application\label{sec:Application}}

Here, we implement the discussions of this paper to practical examples.
First, we consider a simple example of unsupervised learning; a clustering
algorithm based on Shannon entropy minimization. Next we look at some
simple behavior of an intelligent agent as it acts to maximize global
entropy in its environment.

\subsection{Clustering by Entropy Minimization}

Consider a data set consisting of a number of elements organized into
rows. Take for example the data that can be found at \cite{web2}.
This particular example contains 300 samples, each a vector from $\mathbb{R}^{3}$.
This simple proof of concept will group the data into like neighborhoods
by minimizing the entropy across all elements at each respective index
in the data set. This is a data driven example, so essentially we
use a genetic algorithm to perturb the juxtaposition of members of
each neighborhood until the global entropy reaches a minimum (entropic
self organization), while at the same time avoiding trivial cases
such as a neighborhood with only one element.

A prerequisite for running this code is that one must have the Python
framework installed, which is also freely available for many operating
systems at \cite{web3}.

The clustering source code is freely available at \cite{web1}. To
run, simply download it and enter

\begin{lstlisting}
> chmod 777 entropycluster.py
> python entropycluster.py -f entropycluster_data.csv
\end{lstlisting}

Please note that this is a simple prototype, a proof of concept used
to exemplify a concept. It is not optimized for latency, memory utilization,
and it has not been optimized or performance tested against other
algorithms in its comparative class, although dramatic improvements
could be easily acheived by integrating the information content of
the elements into the algorithm. Specifically, we would move elements
with a high information content to clusters where that element would
otherwise have a low information content. Furthermore, observe that
for further efficacy, a preprocessing layer may be beneficial, especially
with topological data sets like the iris data set. Nevertheless, applications
of this concept applied to clustering on small and large scales will
be discussed in a future work.

\begin{figure}
\subfloat[Noisy Data Clustering: Cluster 1.]{

\includegraphics[scale=0.25]{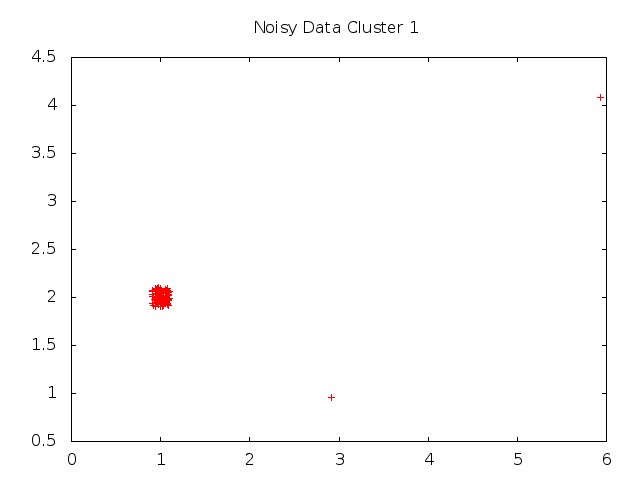}}\hfill{}\subfloat[NoisyData Clustering: Cluster 2.]{\includegraphics[scale=0.25]{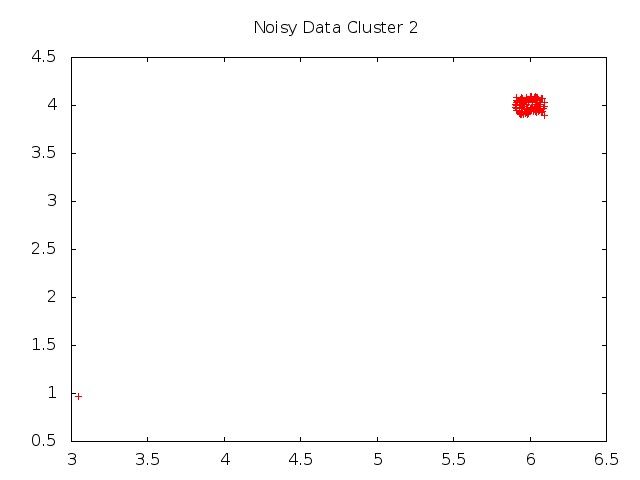}}

\subfloat[NoisyData Clustering: Cluster 3.]{

\includegraphics[scale=0.25]{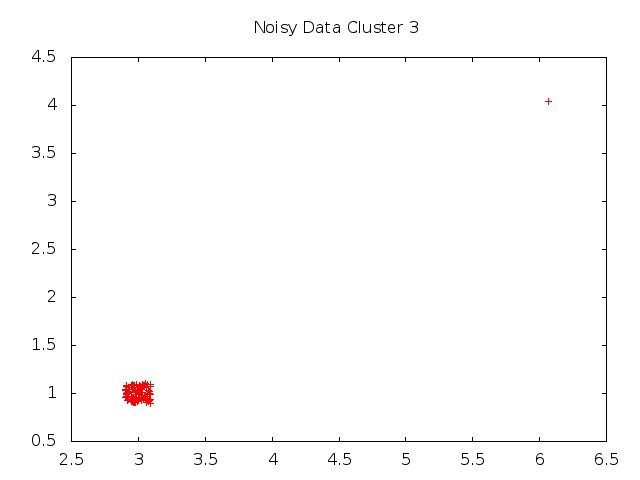}}\protect\caption{Data clustering results with additional noise.\label{fig:noise}}
\end{figure}

We can visualize the progression of the algorithm and the final results,
respectively, in the graphs pictured below. For simplicity, only the
first two (non noise) dimensions are plotted. The accuracy of the
clustering algorithm was 8.3\% error rate in 10000 iterations, with
an average simulation time: 480.1 seconds. Observe that although there
are a few 'blemishes' in the final clustering results, with a proper
choice of parameters including the maximum computational epochs the
clustering algorithm will eventually succeed with 100\% accuracy.
Also pictured in figure \ref{fig:noise} are the results of the clustering
algorithm applied to a data set containing four additional fields
of pseudo-randomly generated noise, each in the interval $[-1,1]$.
The performance of this trial was worse than the last in terms of
speed, but was had about the same classification accuracy. The accuracy
of the clustering algorithm was 6.0\% error rate in 10000 iterations,
with an average simulation time: 1013.1 seconds.

\begin{figure}
\subfloat[Clusters after 2500 iterations.]{

\includegraphics[scale=0.25]{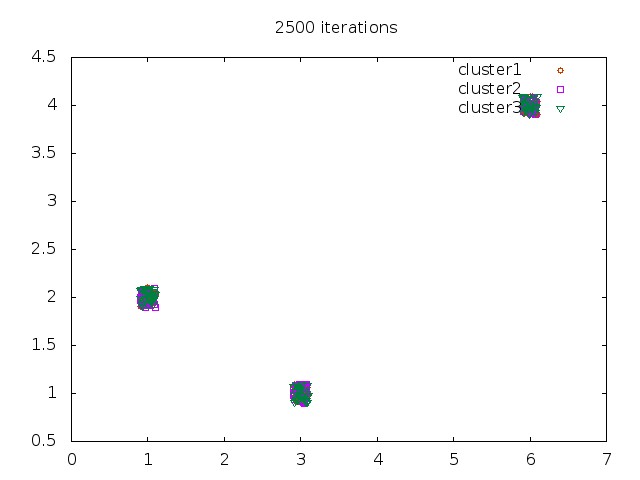}}\hfill{}\subfloat[Clusters after 5000 iterations.]{\includegraphics[scale=0.25]{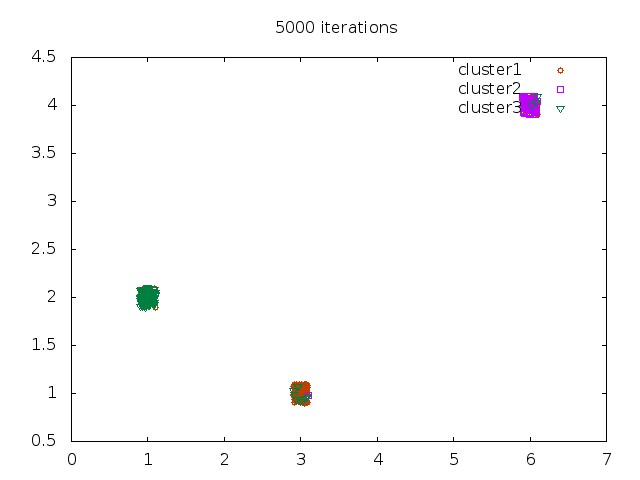}}

\subfloat[Clusters after 7500 iterations]{

\includegraphics[scale=0.25]{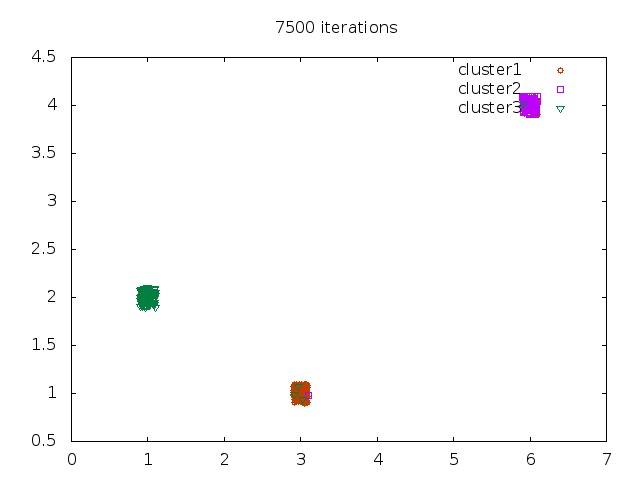}}\hfill{}\subfloat[Clusters after 10000 iterations.]{\includegraphics[scale=0.25]{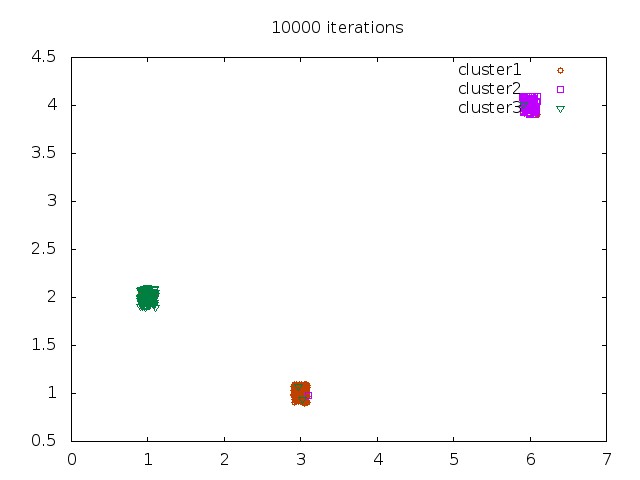}}\protect\caption{Entropic clustering algorithm results over time.}
\end{figure}

\subsection{Global Entropy Maximization}

In our next set of examples consider a virtual agent confined to move
about a 'terrain', represented by a three-dimensional surface, given
by one of the two following equations, respectively:

\begin{equation}
z=\exp[-(x^{2}+y^{2})]\label{eq:f1}
\end{equation}

and

\begin{equation}
z=\frac{1}{4}\exp[-(\left(\frac{x}{10}\right)^{2}+\left(\frac{y}{10}\right)^{2})](\cos[\frac{1}{2}\pi y]+\sin[\frac{1}{2}\pi x]+2).\label{eq:f2}
\end{equation}

We will confine $x,y$ such that $(x,y)\in([x_{\min},x_{\max}],[y_{\min},y_{\max}])$
and note that the range of each respective surface is $z\in[0,1]$.
The algorithm proceeds as follows. First, the agent is initialized
with a starting position, $s=(x_{0},y_{0})$. It updates $s$ by incrementing
or decrementing its coordinates by some small value, $\epsilon=(\epsilon_{x},\epsilon_{y})$.
As the agent meanders about the surface, data is collected as to its
position on the $z-$axis. 

\begin{figure}

\subfloat[Surface as defined in \ref{eq:f1}.]{

\includegraphics[scale=0.25]{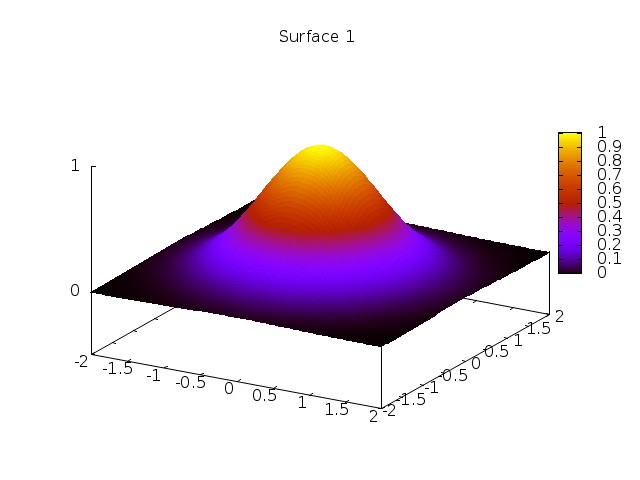}}\hfill{}\subfloat[Surface as defined in \ref{eq:f2}.]{

\includegraphics[scale=0.25]{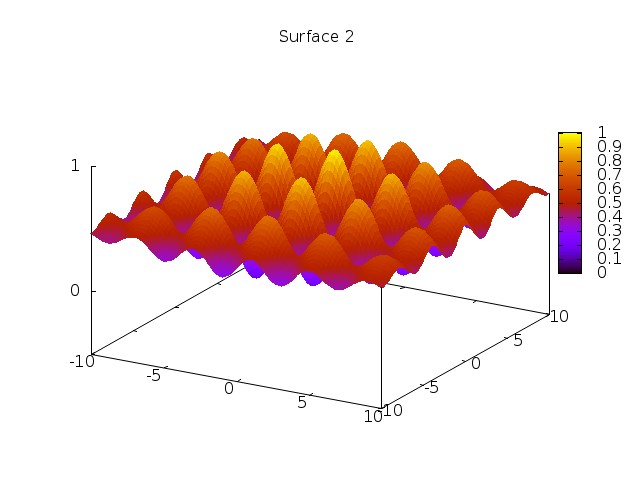}

}\protect\caption{Surfaces for hill climbing agent simulation.}
\end{figure}

If we partition the range of each surface into equally spaced intervals,
we can form a histogram $H$ of the agent's positional information.
From this $H$ we can construct a discrete probability function, $\mathbb{P}_{H}$
and thus calculate the Renyi entropy. The agent can then use feedback
from the entropy determined using $H$ to calculate an appropriate
$\epsilon$ from which it upates its position, and the cycle continues.
The overall goal is to maximize its entropy, or timeout after a predetermined
number of iterations.

In this particular simulation, the agent is initialized using a 'random
walk', in which is $\epsilon$ is chosen at random. Next, it is updated
using feedback from the entropy function.

From the simple set of rules, we see emergent desire for parsimony
with respect to position on the surface, even in the less probable
partitions of $z$, (as$z\rightarrow1$). As our simulation continues
to run, so tends $\mathbb{P}_{H}$ to a uniform distribution.

\begin{figure}
\subfloat[A random walk on surface 2.]{

\includegraphics[scale=0.25]{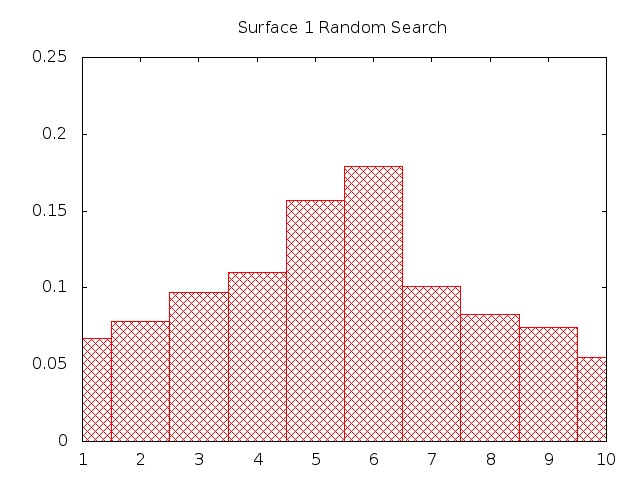}}\hfill{}\subfloat[Continuing to traverse surface 1 using entropic algorithm.]{\includegraphics[scale=0.25]{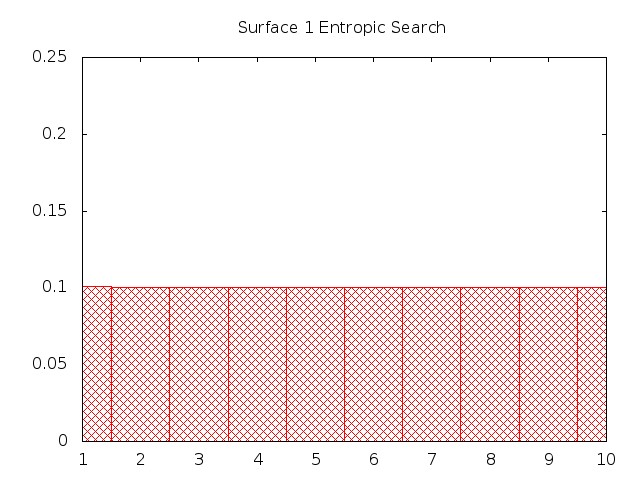}}

\subfloat[Traversing surface 2 using entropic algorithm.]{

\includegraphics[scale=0.25]{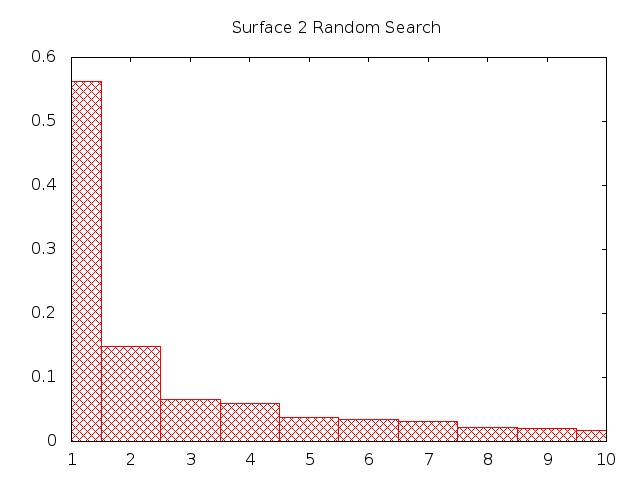}}\hfill{}\subfloat[Continuing to traverse surface 2 using entropic algorithm.]{\includegraphics[scale=0.25]{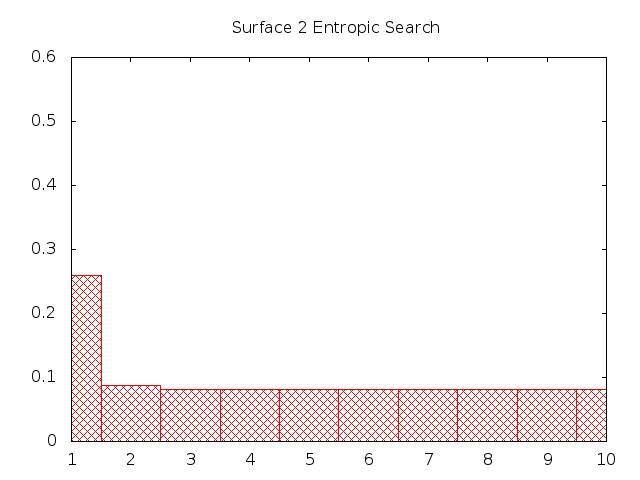}}\protect\caption{Terrain mapping algorithms using random and directed searching.}
\end{figure}
To run the simulation and obtain the above data, simply download the
source code freely available at \cite{web1-1} and enter:

\begin{lstlisting}
> chmod 777 hillclimber.py
> python hillclimber.py
\end{lstlisting}
on the command line.

\section{Related Work\label{sec:Related-Work}}

Although there are many approaches to intelligence from the angle
of cognitive science, few have been proposed from the computational
side. However, as of late, some great work in this area is underway.

Many sources claim to have computational theories of intelligence,
but for the most part these ``theories'' merely act to describe
certain aspects of intelligence \cite{alb}. For example, Meyer in
\cite{meyer} suggests that performance on multiple tasks is dependent
on adaptive executive control, but makes no claim on the emergence
of such characteristics. Others discuss how data is aggregated. This
type of analysis is especially relevant in computer vision and image
recognition \cite{marr}.

The efforts in this paper seek to introduce a much broader theory
of emergence of autonomous goal directed behavior. Similar efforts
are currently under way.

Inspired by physics and cosmology, Wissner-Gross asserts autonomous
agents act to maximize the entropy in their environment \cite{wis}.
Specifically he proposes a path integral formulation from which he
derives a gradient which can be analogized as a causal force propelling
a system along a gradient of maximum entropy over time. Using this
idea, he created a startup called \emph{entropica} that applies this
principal in ingenious ways in a variety of different applications,
ranging from anything to teaching a robot to walk upright, to maximizing
profit potential in the stock market.

Essentially, what Wissner-Gross did was start with a global principal
and worked backwards. What we did in this paper was to was to arrive
at a similar result from a different perspective, namely, entropy
minimization.

\section{Conclusion\label{sec:Conclusion}}

The purpose of this paper was to lay the groundwork for a generalization
of the concept of intelligence in the computational sense. We discussed
how entropy minimization can be utilized to facilitate the intelligence
process, and how the disparities between the agent's prediction and
the reality of the training set can be used to optimize the agents
performance. We also showed how such a concept could be used to produce
a meaningful, albeit simplified, practical demonstration.

Some future work includes applying the principals of this paper to
data analysis, specifically in the presence of noise or sparse data.
We will discuss some of these applications in the next paper.

More future work includes discussing the underlying principals under
which data can be collected hierarchically, discussing how computational
processes can implement the discussions in this paper to evolve and
work together to form processes of greater complexity, and discussing
the relevance of these contributions to abstract concepts like consciousness
and self awareness.

In the following paper we will examine how information can aggregate
together to form more complicated structures, the roles these structures
can play. More concepts, examples, and applications will follow in
future works.

\end{document}